\documentclass[11pt]{article}

\usepackage[margin=1in]{geometry}
\usepackage[utf8]{inputenc}
\usepackage[T1]{fontenc}
\usepackage{lmodern}
\usepackage{microtype}
\usepackage{graphicx}
\usepackage{booktabs}
\usepackage{multirow}
\usepackage{amsmath,amssymb}
\usepackage[numbers,sort&compress]{natbib}
\usepackage{tikz}
\usetikzlibrary{arrows.meta,positioning,fit,backgrounds}
\usepackage{xcolor}
\usepackage[colorlinks=true,linkcolor=blue!60!black,citecolor=blue!60!black,urlcolor=blue!60!black]{hyperref}
\usepackage{orcidlink}
\usepackage{caption}
\newcommand{\Fmax}{\ensuremath{F_{\max}}}
\newcommand{\Fbreak}{\ensuremath{F_{\mathrm{break}}}}
\newcommand{\Fcmd}{\ensuremath{F_{\mathrm{cmd}}}}
\newcommand{\code}[1]{\texttt{\small #1}}

\title{\textbf{FORGE-plus: Force-Budgeted Recovery for Contact-Rich Assembly\\ with a Frozen LLM Supervisor}\\[6pt]
\large A simulation study of who sets the force ceiling, and what to do on failure}

\author{%
Kyupaeck Jeff Rah\,\orcidlink{0000-0003-1898-2930}\\
\normalsize Independent Researcher\\
\normalsize\texttt{jeffrah89@gmail.com}
\and
Midum Oh\\
\normalsize Independent Researcher\\
\normalsize\texttt{midumoh@gmail.com}}

\date{July 2026}

\begin{document}
\maketitle

\begin{abstract}
Force-conditioned RL skills can seat tight-clearance assemblies while respecting a commanded force ceiling, but two questions remain open: \emph{who sets the ceiling}, and \emph{what should the robot do when an insertion fails}. Autotuning the ceiling for success works only for indestructible parts, and the natural recovery --- press harder --- is precisely the action that destroys a fragile one. We study a two-layer system in which a \textbf{frozen, text-only LLM} (a) sets a per-object force ceiling \Fmax{} from the object's identity before the episode, and (b) on failure, selects a recovery maneuver from a fixed menu by reading a \textbf{compact text force signature} --- no images anywhere in the system. Force authority never belongs to the LLM: a hard clamp in the fast control loop saturates commanded forces to \Fmax{}, the recovery menu contains no ``increase the ceiling'' option, and the hidden per-instance breaking force \Fbreak{} is visible only to the evaluator, making the fragility metric non-circular. On a fragile bottle-into-rack placement and a 0.4\,mm-diametral-clearance gear insertion, each executed on two grippers (Robotiq 2F-140 and Franka Panda hand), a single unified checkpoint passes strict 256/256 clean gates on both a fragile and a robust object class with zero breaks, decides its own release via a learned head (256/256, zero bad releases), and completes a fully physical table-pick flow at a 5.4\,N mean peak force. Under an injected in-grip slip, the force-signature chain recovers 40\% (2F-140) and 64\% (Franka) of jams, while a press-harder baseline shows two distinct failure faces: it is 100\% futile on one gripper and breaks 96\% of fragile parts on the other. We report our negative results as first-class findings: PPO provably fails at this clearance on the 2F-140 (exploration noise that can find the funnel already breaks the part), tiny-std PPO polish destroys a working policy, and three natural designs for a learned release head fail for instructive reasons. All experiments are in rigid-body simulation with breakage modeled as a hidden scalar force threshold; no sim-to-real claim is made. Videos, code, and evaluation logs: \url{https://robot-team00.github.io/FORGE-plus/}.
\end{abstract}

\section{Introduction}
\label{sec:intro}

The Factory$\rightarrow$IndustReal$\rightarrow$AutoMate$\rightarrow$FORGE line~\citep{narang2022factory,tang2023industreal,tang2024automate,noseworthy2025forge} has made simulated contact-rich insertion robust and, in several cases, transferable to hardware. FORGE~\citep{noseworthy2025forge} in particular conditions its policy on a maximum allowable force, so the skill can trade speed against gentleness. But that progress leaves two questions unanswered.

\textbf{Who sets the ceiling?} FORGE autotunes \Fmax{} (or takes it from a human) to maximize success. That is the right choice for indestructible parts and the wrong one the moment a single line mixes a steel bolt with a nylon snap clip: the ceiling that seats the bolt strips the clip. The budget must be \emph{per object}, and it has to come from somewhere --- most naturally from what the object \emph{is}.

\textbf{What do you do on failure?} Existing LLM/VLM failure reasoners --- REFLECT, DoReMi, AHA~\citep{liu2023reflect,guo2023doremi,duan2024aha} --- read vision and re-plan at the task level. But the information that distinguishes a wedge from a cross-thread from a burr lives in the \emph{force trace}, not the pixels: those failures often look identical on camera and feel completely different. And the obvious recovery --- press harder, a published behavior in Tactile-VLA~\citep{huang2025tactilevla} --- is exactly the move that finishes a robust part and destroys a fragile one.

FORGE-plus closes both gaps at once with a deliberately thin semantic layer:
\begin{itemize}
  \item A \textbf{frozen LLM} reads the object's identity (text only, no images) and sets a per-object \Fmax{} before the episode starts.
  \item On failure, the same LLM reads a \textbf{compact force/contact signature} and picks a recovery from a fixed menu that has no ``increase the ceiling'' option.
  \item \textbf{Safety is enforced by a hard clamp in the fast control loop} --- never by the language model. The LLM emits a number and a menu choice; the clamp decides what force is physically commanded.
\end{itemize}

Three invariants make the evaluation non-circular (\S\ref{sec:invariants}): the hidden breaking force \Fbreak{} is never observed by any learned or LLM component; \Fmax{} is immutable during recovery; and force authority lives in the fast loop. The single way to score well on breakage is to set and respect a sensible ceiling from what the part \emph{is}.

\paragraph{Scope.} This is a \textbf{simulation-only} study of the terminal ``micro'' phase of manipulation --- the last few centimeters of contact-rich seating or insertion, where force decides success and breakage. No component uses vision: the skill observes proprioception, pose, and force/torque; the LLM reads text. Object identity is operator-provided and the poses of part and target are taken from privileged simulator state; perception, part identification, and long-range transport are upstream problems, deliberately out of scope. Breakage is modeled as a hidden scalar threshold on peak contact force --- there is no fracture or deformation modeling --- and no sim-to-real transfer is claimed.

\paragraph{Contributions.}
\begin{enumerate}
  \item \textbf{A two-layer architecture and benchmark} for force-budgeted assembly under per-object fragility, with a non-circular breakage metric (hidden per-episode \Fbreak{} draws), seven evaluation metrics, and six baselines including a press-harder recovery modeled on Tactile-VLA's published behavior (\S\ref{sec:system}, \S\ref{sec:benchmark}).
  \item \textbf{Full-cycle results on two grippers.} On the Robotiq 2F-140, one unified checkpoint passes strict clean gates of 256/256 on both a fragile ABS gear and a steel gear with zero breaks, a learned release head passes 256/256 with zero bad releases, and a fully physical table-pick flow completes 64/64 at a 5.4\,N mean peak insertion force --- the gentlest contact of the project (\S\ref{sec:exp-gates}).
  \item \textbf{Force-signature recovery vs.\ baselines.} On the bottle task, a rim wedge is caught from the force signature at 16.1\,N against a 23\,N break, and budget-preserving retries seat the fragile bottle in 9/9 headless recovery episodes with zero breaks (\S\ref{sec:exp-task3}). On the gear task, under a 5\,mm in-grip slip, only the force-signature chain ever routes to \code{regrasp} --- the one maneuver that fixes a tilted grip --- recovering 40\% (2F-140) / 64\% (Franka) of jams vs.\ 28\% / 32\% for a menu-sampling proxy of a vision-based reasoner and 0\% for a hand-coded heuristic. Press-harder shows \emph{two different failure faces} on the two grippers: futile (0\% success, 0 breaks, 100\% timeouts) on the 2F-140, destructive (96\% breaks) on the Franka (\S\ref{sec:exp-recovery}).
  \item \textbf{Negative results as findings.} PPO cannot learn this insertion on the 2F-140 at 0.4\,mm clearance --- the exploration noise needed to search the funnel already breaks the part (nine escalating runs, zero seats) --- and tiny-std PPO polish destroys a working BC policy within 25 iterations. Three natural designs for a learned release head fail for reasons that generalize (\S\ref{sec:skill}). The oracle budget \Fbreak$-\varepsilon$ breaks half the fragile parts because command clamping does not bound contact overshoot (\S\ref{sec:exp-budget}).
\end{enumerate}

\section{Related work}
\label{sec:related}

\textbf{The skill: FORGE.} We reuse FORGE-style force conditioning~\citep{noseworthy2025forge} essentially wholesale: the policy takes a normalized force command as an observation, is penalized for overshooting it, and learns to seat under the ceiling. The difference is upstream --- FORGE chooses \Fmax{} to make the task succeed; we choose it to keep the part alive, from identity, and the recovery layer is forbidden from relaxing it.

\textbf{The closest semantics: Tactile-VLA.} Tactile-VLA~\citep{huang2025tactilevla} performs language-conditioned force control and demonstrates ``reasoning to overcome failure'' by autonomously \emph{increasing} force. That behavior is precisely our press-harder baseline. We differ in using a frozen LLM rather than a fine-tuned VLA, a force-only signature with no vision, a hard per-object ceiling recovery cannot cross, and an evaluation containing fragile parts that press-harder destroys. In Tactile-VLA's setting there is no breakage to violate, so force escalation is safe by construction; ours is the setting where it is not.

\textbf{The closest safety architecture: PaCo-VLA.} PaCo-VLA~\citep{cao2026pacovla} shares the commitment that the network proposes while a runtime shield retains authority; its shield is a passivity/energy-tank contract over admittance updates, explicitly not a bound on peak force. Ours is exactly a hard scalar peak-force clamp tied to object identity. ForceVLA~\citep{yu2025forcevla} and CompliantVLA-adaptor~\citep{zhang2026compliantvla} provide further force-aware VLA context.

\textbf{Failure reasoners: REFLECT, DoReMi, AHA.} All three~\citep{liu2023reflect,guo2023doremi,duan2024aha} are vision-centric and operate at the task-plan level (wrong object, missing step, mis-grasp). Our failure signal is a compact force/contact signature, the recovery is constrained by a force budget, and the failures we target live in contact dynamics that look alike on camera.

\textbf{Assembly benchmarks.} Factory~\citep{narang2022factory}, IndustReal~\citep{tang2023industreal}, and AutoMate~\citep{tang2024automate} established the simulation line we build on (we use Isaac Lab~\citep{mittal2025isaaclab,mittal2023orbit}); none models per-object fragility or force-grounded recovery. Grasping embodiments follow GraspGen~\citep{murali2025graspgen}, whose two parallel-jaw grippers (Franka Panda hand, Robotiq 2F-140) are our generalization axis.

No single ingredient here is new; the contribution is the conjunction --- force-grounded (not vision) semantic recovery, under a hard per-object ceiling the recovery cannot raise, scored closed-loop with a fragility metric the agent cannot read off.

\section{System}
\label{sec:system}

\begin{figure}[t]
\centering
\resizebox{\textwidth}{!}{%
\begin{tikzpicture}[
  font=\small,
  box/.style={draw, rounded corners=2pt, align=center, inner sep=5pt, minimum height=9mm, fill=white},
  slow/.style={box, draw=blue!60!black, thick},
  fast/.style={box, draw=black!70, thick},
  guard/.style={box, draw=orange!80!black, thick},
  arr/.style={-{Stealth[length=2.2mm]}, thick, blue!60!black},
  arrs/.style={-{Stealth[length=2.2mm]}, thick, black!70},
  lab/.style={font=\scriptsize, inner sep=1pt}
]
% slow layer
\node[slow] (budget) {Budget-Setter\\ \scriptsize identity $\to$ \Fmax};
\node[slow, right=9mm of budget] (recov) {Recovery-Selector\\ \scriptsize signature $\to$ menu action};
\node[slow, right=9mm of recov, fill=blue!4] (enc) {Signature Encoder\\ \scriptsize F/T history $\to$ text\\ \scriptsize \textcolor{red!70!black}{no images $\cdot$ no \Fbreak}};
\begin{scope}[on background layer]
\node[draw=blue!40, fill=blue!2, rounded corners=3pt, fit=(budget)(recov)(enc), inner sep=7pt, label={[lab, blue!60!black]north west:SLOW LAYER --- frozen LLM ($\sim$0.1--1 Hz)}] (slowband) {};
\end{scope}
% fast layer
\node[fast, below=17mm of budget] (skill) {RL skill\\ \scriptsize conditioned on \Fcmd};
\node[fast, right=7mm of skill] (ctrl) {Controller\\ \scriptsize impedance / OSC};
\node[guard, right=7mm of ctrl] (clamp) {\textbf{Force clamp}\\ \scriptsize saturate to \Fmax\\ \scriptsize \textcolor{red!70!black}{safety $\neq$ the LLM}};
\node[fast, right=7mm of clamp] (robot) {Robot + object\\ \scriptsize Isaac Lab, rigid body};
\begin{scope}[on background layer]
\node[draw=black!40, fill=black!3, rounded corners=3pt, fit=(skill)(ctrl)(clamp)(robot), inner sep=7pt, label={[lab, black!70]north west:FAST LAYER ($\sim$60--120 Hz) --- force authority lives here}] (fastband) {};
\end{scope}
% evaluator
\node[guard, dashed, below=8mm of robot, xshift=-9mm] (eval) {Evaluator only: hidden \Fbreak{} per instance\\ \scriptsize breakage $\iff$ peak contact $>$ \Fbreak{} --- never an input to policy or LLM};
% arrows
\draw[arr] (budget.south) -- node[lab, left, blue!60!black]{\Fmax} (skill.north);
\draw[arr] (budget.south east) to[out=-60,in=120] (clamp.north west);
\draw[arr] (recov.south) to[out=-90,in=60] node[lab, left, pos=0.6, blue!60!black]{recovery action} (skill.north east);
\draw[arrs] (robot.north) to[out=90,in=-90] node[lab, right, pos=0.45]{F/T, contact} (enc.south);
\draw[arr] (enc.west) -- node[lab, above]{signature} (recov.east);
\draw[arrs, dashed, orange!80!black] (eval.north) to[out=90,in=-90] (robot.south);
\end{tikzpicture}%
}
\caption{The two-layer architecture. The LLM only emits a number (\Fmax{}) and a menu choice; the guarantee that commanded force never exceeds \Fmax{} is produced by the clamp in the fast loop, backed by a fixed 120\,N global hard cap and JSON range validation. The hidden \Fbreak{} feeds only the evaluator's breakage check.}
\label{fig:arch}
\end{figure}
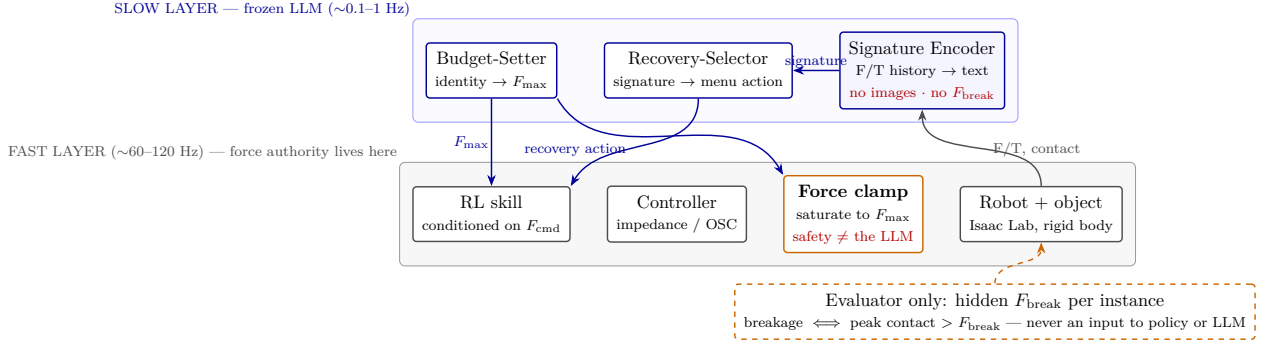

Two layers run at different rates and never blur their roles (Figure~\ref{fig:arch}). The \textbf{slow layer} is a frozen LLM called at most twice per episode --- once at episode start to set the budget, once per failure to pick a recovery --- with JSON in and JSON out, no images. The \textbf{fast layer} is the force-conditioned RL skill plus an operational-space impedance controller at control rate, where force authority lives. A \emph{signature encoder} between them turns the recent force/torque and contact history into a short text token.

\subsection{The two LLM calls}

\textbf{Budget-setter.} Input: the object's identity --- name, material, class, nominal mass, geometry tags (e.g.\ \code{thin\_wall}, \code{press\_fit}) --- plus the task and the global hard cap. Output: a numeric \code{F\_max\_N} (optionally per-axis), a confidence, and a rationale. The output is range-validated and clamped to a fixed global hard cap (120\,N) before it reaches the controller; budget-setter outputs are cached per object class since the call is near-deterministic. The LLM reasons about the \emph{class} --- it can never memorize an instance, because \Fbreak{} is sampled per episode (\S\ref{sec:benchmark}).

\textbf{Recovery-selector.} Input: the current \Fmax{} (echoed read-only), the attempt count, and the force signature:
\begin{center}
\code{\{peak\_axial\_N, net\_insert\_mm, axial\_rising, lateral\_bias, contact\_persist\_ms, slip\_events\}}
\end{center}
Output: one action from the fixed menu
\code{\{retract\_and\_reapproach, wiggle\_search, rotate\_align, regrasp, abort\}}
with bounded parameters. The menu deliberately omits any ``increase the ceiling'' option, and the returned \code{keep\_F\_max\_N} is overwritten to the original value server-side regardless of what the model returns. Recovery reallocates \emph{motion}, not force. The recovery \emph{primitives} are fixed scripted maneuvers; the LLM's contribution is the \emph{selection}, grounded in the signature.

The LLM backend is swappable (hosted API, a local 7--8B instruct model, or a deterministic force-reasoned heuristic for CI); all backends read only text.

\subsection{Design invariants (the non-circularity of the evaluation)}
\label{sec:invariants}

\begin{enumerate}
  \item \textbf{\Fbreak{} is never observed by the agent.} It is sampled once per episode from the object class's distribution, stored in the simulator, and read only by the evaluator's breakage check. The policy, the signature encoder, the budget-setter, and the recovery-selector never see it; a runtime assertion in \code{SignatureEncoder.encode()} guards this at test time.
  \item \textbf{\Fmax{} is immutable during recovery.} The recovery-selector's \code{keep\_F\_max\_N} is always overwritten back to the original value before it reaches the controller.
  \item \textbf{Force authority is in the fast loop.} The \code{ForceClamp} saturates the commanded wrench before every simulation step regardless of what the LLM said; a fixed 120\,N global cap is a second line of defense against hallucinated ceilings.
\end{enumerate}

One subtlety we do not hide: clamping the \emph{command} does not by itself bound the \emph{contact} force --- impedance overshoot can exceed \Fmax{}. We mitigate with low controller stiffness and bounded per-step motion, and we measure the gap as a first-class metric (clamp fidelity). Section~\ref{sec:exp-budget} shows this is not a technicality: it is why the oracle budget breaks parts.

\subsection{The closed loop}

Each episode runs the skill until success or a detected failure; on failure the signature is encoded, the LLM picks a recovery, the fixed primitive executes within the same \Fmax{}, and the skill resumes --- up to $K_{\max}$ attempts. Failure detection is deliberately conservative and force-grounded: a \emph{jam} is sustained contact with no net descent over a window (a wedge, not a force spike --- a clean insertion grind that does descend passes through), and a \emph{contactless hover} branch catches the policy correctly refusing a geometrically impossible insertion (\S\ref{sec:exp-recovery}). A false positive costs one benign lift-and-retry attempt, never more force.

\section{Benchmark}
\label{sec:benchmark}

\subsection{Tasks, objects, embodiments}

\textbf{Bottle placement --- a fragile bottle into a wine rack.} The robot inserts a LIBERO wine bottle, held by the neck under real friction, into a cell of a 3$\times$3 rack, then releases it standing upright --- a place-and-release peg-in-hole where ``press harder'' is maximally destructive. Fragile class: \Fbreak\,$\approx$\,22\,N, budget 8.8\,N; robust class: 180\,N. (The rendered mesh is the wine bottle in all cases; the class sets the hidden fragility.) The recovery episode is demonstrated on both grippers (Robotiq 2F-140 and Franka Panda hand); the quantitative arc is on the Franka.

\textbf{Gear insertion --- tight clearance.} The FORGE GearMesh medium gear ($\varnothing$35.5\,mm hub, bore fit band $r$\,=\,5.15\,mm, \textbf{0.4\,mm diametral clearance}) must be inserted onto the middle shaft of a three-shaft base plate. Two object classes share the geometry: \code{abs\_gear} (fragile; \Fbreak{} drawn per episode from $38\pm5$\,N; identity-derived budget $\approx$10\,N insertion) and \code{steel\_gear} (robust; 100\,N budget). Success is a \emph{strict true seat}: gear origin within 0.6\,mm of the seated height, upright, on the correct shaft. The full quantitative arc runs on \textbf{both grippers}: the Robotiq 2F-140 four-bar adaptive gripper and the Franka Panda hand.\footnote{The 2F-140 port required building a combined Franka+2F-140 USD asset (NVIDIA ships that combination only for the 2F-85) and two PhysX-level findings documented in the repo: the four-bar loop joints do not survive articulation teleports (hence a no-teleport contract: spawn at the authored pose, drive only by position targets), and the merged gripper's excluded loop joints materialize only when a standalone instance of the same gripper USD exists in the scene.}

\textbf{Disturbances.} Bottle placement injects a lateral base-aim error that wedges the bottle on the cell rim. Gear insertion injects a \textbf{5\,mm in-grip slip}: the gear is displaced in the pinch mid-carry. The pads re-center the \emph{position} within a substep, but the \emph{orientation} does not recover --- the asymmetric squeeze leaves the gear tilted 7--10$^\circ$ in the grip, which cannot thread a 0.4\,mm-clearance bore ($\sim$3--4$^\circ$ max while threaded).

\subsection{What is learned vs.\ what is scripted}
\label{sec:learned-scripted}

The split is explicit, and the released videos label every phase on-screen in real time (green \textsc{learned} / orange \textsc{scripted}):

\emph{Learned}: the force-guided insertion --- funnel search, alignment, descent, seating press, everything from the entrance hand-off to the seat --- and the \emph{release decision} (an 8th action dimension; the environment opens the gripper only when the learned head crosses its threshold).

\emph{Scripted}: pre-approach transport between known poses (episode staging, as in the underlying benchmarks), the recovery \emph{primitives} (the menu the LLM selects from), and the post-release hand retract. In the final table-pick flow, nothing about the object is teleported, pinned, or kinematically attached at any point: the pick closes a real friction grip on the resting gear, and everything after is genuine physics.

\subsection{Metrics and baselines}

Seven metrics: (1) closed-loop multi-attempt success; (2) \textbf{breakage rate} (peak contact $>$ \Fbreak{}; non-circular since \Fbreak{} is evaluator-only); (3) budget appropriateness (the margin $m=\Fbreak-\Fmax$, over-/under-budget rates); (4) recovery efficacy; (5) force economy (peak/mean contact per success); (6) clamp fidelity (contact overshoot above \Fmax); (7) cross-gripper transfer.

Six baselines: \emph{no-ceiling}, \emph{fixed global ceiling}, \emph{press-harder} (per-object budget, recovery escalates force $\times$1.25 per attempt --- Tactile-VLA's published recovery behavior), \emph{heuristic recovery} (hand-coded rules on the same signature), \emph{vision-LLM proxy} (a menu-sampling stand-in for pixels-only reasoners --- see the caveat in \S\ref{sec:exp-recovery}), and the \emph{oracle ceiling} ($\Fbreak-\varepsilon$; cheats by reading the hidden variable).

\section{Skill learning: what works at 0.4\,mm, and what provably does not}
\label{sec:skill}

The skill is a force-conditioned MLP policy (34-dim observation: proprioception, EE pose, F/T, object-up axis, phase; \Fcmd{} conditions the network) whose position deltas ride on an operational-space impedance controller with bounded per-step motion. On the Franka hand, FORGE-style PPO~\citep{schulman2017ppo} training suffices (clean gate: 200/200 strict seats, 0 breaks). Porting to the Robotiq 2F-140 exposed three plant-level defects that made every policy fail before learning was in question --- the four-bar's paired knuckles must be driven symmetrically, the wrist target must be anchored and command-continuous at hand-off, and the funnel search needs lateral authority while in contact --- and then a clean negative result:

\subsection{PPO provably fails at this clearance (negative result 1)}
\label{sec:ppo-fails}

The funnel tolerance ($\sim$1.5\,mm capture radius; 0.25\,mm effective bore clearance at $\mu$\,=\,1.0) sits far below workable exploration noise: an action std of 0.12 already breaks 40/64 fragile gears while seating 1/320. \textbf{Nine escalating PPO runs on the 2F-140 --- spanning plant fixes, noise schedules, annealing, and contact-observation variants --- never seated a single gear}; all nine checkpoints are kept in the repository manifest as evidence. The exploration noise needed to \emph{search} the funnel is already destructive at the funnel's own scale: for fragile tight-clearance assembly, the exploration--damage trade-off can exclude on-policy RL outright.

A second, sharper negative: \textbf{tiny-std PPO polish destroys a working policy}. Fine-tuning a functioning BC policy with PPO at std 0.05 collapsed it from 84\% to 0/32 within 25 iterations, at every snapshot. The mechanism is the $1/\sigma^2$ term in the Gaussian policy gradient: as std shrinks, likelihood-ratio gradients are amplified so that the mean moves far more coarsely than the 0.4\,mm task tolerates.

\subsection{What works: demos $\to$ DAgger $\to$ BC $\to$ weight soup}

Scripted experts are permitted for \emph{training data only} (never in evaluation). The pipeline: collect expert demonstrations, run DAgger~\citep{ross2011dagger} rounds relabeling the policy's own visited states, and fit the deterministic mean by behavior cloning. The lineage climbs from closed-loop failure (plain BC: 30/32 broke) through DAgger rounds to \code{bc3} (fragile class 256/256) and \code{bc7} (best single: steel 32/32, fragile 84\%). The two classes tug the policy in opposite directions --- the fragile class sits on its own break floor (\Fbreak{} draws as low as $\sim$20\,N vs.\ 15--25\,N contact transients) --- and balanced-refit attempts see-sawed.

\textbf{Unification is a weight soup}~\citep{wortsman2022soups}: \code{uni} $= 0.15\cdot$\code{bc3}$+0.85\cdot$\code{bc7}. The interpolation is valid because \code{bc7} was warm-started from \code{bc3} (same basin); every $\alpha\in[0.10,0.75]$ passes triage, steel peak force is monotone in $\alpha$, and $\alpha=0.15$ minimizes the joint force tails. One checkpoint then passes both clean gates (\S\ref{sec:exp-gates}).

\subsection{The learned release head (negative results 2--4, and the winner)}
\label{sec:release}

The policy should decide \emph{when to let go} --- an 8th action dimension --- without disturbing the proven arm behavior. Three natural designs fail for general reasons:
\begin{enumerate}
  \item \emph{Frozen-trunk head trained on all timesteps} learns ``open iff already open'' and never fires at deployment. The environment latches the first release, so every post-open observation carries no decision information --- \textbf{post-open samples are label poison}; drop them always.
  \item \emph{Pre-open-only labels on the frozen trunk} yields 27\% false positives / 52\% false negatives: the arm-trained trunk provably \emph{discards} the seat-state signal, having never needed it.
  \item \emph{Trunk fine-tuning with an arm-distillation loss} sees the two losses fight; the false-negative rate pins at 50\% with zero usable coverage.
\end{enumerate}
\textbf{The winner is an input-skip head}: a single linear layer reading the \emph{raw observation} (plus \Fcmd), bypassing the trunk. A logistic-regression diagnostic first proved the release signal is linearly separable in the raw observation (zero false negatives, zero-false-positive window coverage 255/256), so only that linear layer is trained --- the arm dimensions and trunk stay bit-identical to \code{uni}. Two deployment traps with teeth: the feature standardization must be folded into the weights and the operating threshold set \emph{after} folding (fp32 fold error scales with logit magnitude); and the collection-time settle matters --- labeling release 20 steps after the geometric seat let the seating press continue and \emph{broke 36\% of fragile gears before the release could fire}; a 5-step settle gives 256/256. The head reads a phase one-hot, so it is staging-specific and must be recollected per deployment flow.

\subsection{Physical staging: the table pick and the place-on-table regrasp}
\label{sec:staging}

The last non-physical assist in the pipeline was a seat-window pin that placed the gear in the closed gripper at episode start. The final flow replaces it end-to-end: the gear spawns resting on the table, the gripper descends open, closes a real friction grip on the hub, lifts, carries, and hands off to the policy. Two durable lessons: precise transports must run on stiff joint drives, not the operational-space controller --- an OSC wrist-twist disturbance walked the end-effector 15--30\,mm sideways during any settle or traverse, and this same defect independently killed the carry, the hover, and the recovery place-traverse before being rooted out --- and grip geometry tolerances are load-bearing (pads 6\,mm low on the hub produce a 3.5$^\circ$ carry lean, which the policy correctly \emph{refuses} to insert). In recovery, \code{regrasp} becomes fully physical: place the tilted gear back on the table, open, re-pick it upright, carry back, resume. Safety rules found by root-causing (each with a recorded incident): drives may engage only in confirmed free space ($<$0.5\,N contact; the timeout-to-drives path once crushed a wedged gear at 213\,N), every drive engage must droop-compensate ($\sim$3.5\,cm sag under load), and the place descent must gate on the realized gear height, not the arm pose (a held gear rides $\sim$30\,mm lower than a resting one; 319\,N observed otherwise).

\section{Experiments}
\label{sec:exp}

All results are in Isaac Lab rigid-body simulation with the deterministic policy mean, strict success criteria, and hidden per-episode \Fbreak{} draws. Every number below is reproducible from the repository's evaluation scripts and is stated with its protocol.

\subsection{Bottle placement: a jam caught at 16.1\,N against a 23\,N break}
\label{sec:exp-task3}

\begin{figure}[t]
\centering
\includegraphics[width=0.55\textwidth]{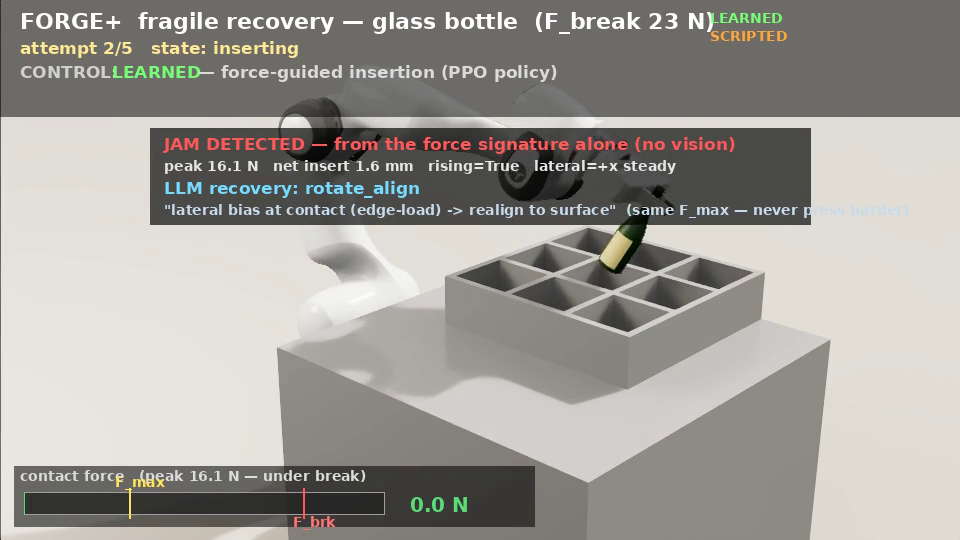}
\caption{The rendered fragile recovery episode (Franka, glass-class bottle, \Fbreak\,$\approx$\,23\,N). The JAM card shows the actual signature the LLM saw --- peak 16.1\,N, net insert 1.6\,mm, rising, lateral bias \code{+x steady} --- its \code{rotate\_align} decision, and the standing constraints: \emph{from the force signature alone (no vision)}; \emph{same \Fmax{} --- never press harder}.}
\label{fig:franka-jam}
\end{figure}

The recovery episode is rendered on both grippers. On the Robotiq 2F-140, a seeded base-aim fault wedges the fragile bottle on the rack; the jam is caught from the force signature (peak 13.9\,N against that episode's 19\,N break), and the second attempt places the bottle upright. The quantitative arc is on the Franka: a FORGE-style PPO policy (8-dim action including the learned release) performs the insertion under gentle contact ($\sim$0--4\,N against the fragile budget) at $\approx$68\% single-attempt success; the recovery loop is what makes it reliable. With an induced rim wedge on the \emph{fragile} class, the jam is caught from the force signature at \textbf{16.1\,N --- far below the episode's 23.3\,N break} --- the LLM picks \code{rotate\_align}, and the learned policy re-descends and seats the bottle: \textbf{9/9 headless recovery episodes seat the fragile bottle} (2--3 attempts each, 0 breaks), even though a single attempt under the low budget rarely seats it. The same loop on the robust class catches the wedge at $\sim$17\,N vs.\ a 180\,N break. The honest value of the layer is exactly this: the conservative detector plus benign, budget-preserving retries turn an imperfect learned insertion into a reliable one, whether the stall came from the injected misalignment or the policy's own marginality. (The rendered video's stricter full-place gate --- upright settle after a learned release the head was never trained to time on this object class --- passes in about 1 of 3 takes; the release dimension is gated off during recovery and sampled at the seat, as documented in the repo.)

\subsection{Clean gates: one checkpoint, both classes, zero breaks}
\label{sec:exp-gates}

\begin{table}[t]
\centering
\small
\caption{Robotiq 2F-140 clean gates (deterministic policy, strict true-seat criterion, hidden per-episode \Fbreak{}; 256 episodes per class unless noted). Release success is stricter than insertion success: released + gear standing seated + hand retracted clear, with the environment opening the gripper only on the learned head's own threshold crossing.}
\label{tab:gates}
\resizebox{\textwidth}{!}{%
\begin{tabular}{llll}
\toprule
Milestone & Checkpoint & Fragile ABS gear & Steel gear \\
\midrule
Clean insertion (unified) & \code{uni} & \textbf{256/256, 0 breaks}; peak 15.9\,N mean / 18.5 p95 & \textbf{256/256, 0 breaks}; 35.6 / 57.6 \\
+ learned release & \code{uni\_rel} & \textbf{256/256, 0 breaks, 0 bad releases}; 15.7 / 18.5 / 20.1 max & \textbf{256/256, 0, 0}; 35.7 / 59.3 / 69.6 \\
Full table-pick flow & \code{uni\_rel\_tp} & \textbf{64/64 (smoke), 0 breaks, 0 bad rel.}; peak \textbf{5.4\,N mean} / 5.8 max & --- \\
\bottomrule
\end{tabular}%
}
\end{table}

Table~\ref{tab:gates} summarizes the Robotiq 2F-140 arc; the Franka clean gate is 200/200 strict seats, 0 breaks (peak 13.8\,N mean / 15.9 p95 / 17.6 max). The unified checkpoint records zero over-budget episodes in both classes.

Two observations. First, the weight-soup unification is not cosmetic: no single BC checkpoint passed both class gates, and PPO could not train one (\S\ref{sec:ppo-fails}). Second, an instructive inversion: the fully physical \emph{table-pick} flow produces far \emph{gentler} insertions (5.4\,N mean peak) than the pinned staging it replaced (15.9\,N). The staging pin left a small systematic grip offset that the policy paid for during the funnel search; a real friction pick centers the hub in the pads better than the pin ever did. Removing the last non-physical assist made the force economy better, not worse. (The 64-episode table-flow gate is a smoke, not a full 256-episode gate; the steel-class table-pick smoke is open work.)

\begin{figure}[t]
\centering
\includegraphics[width=0.24\textwidth]{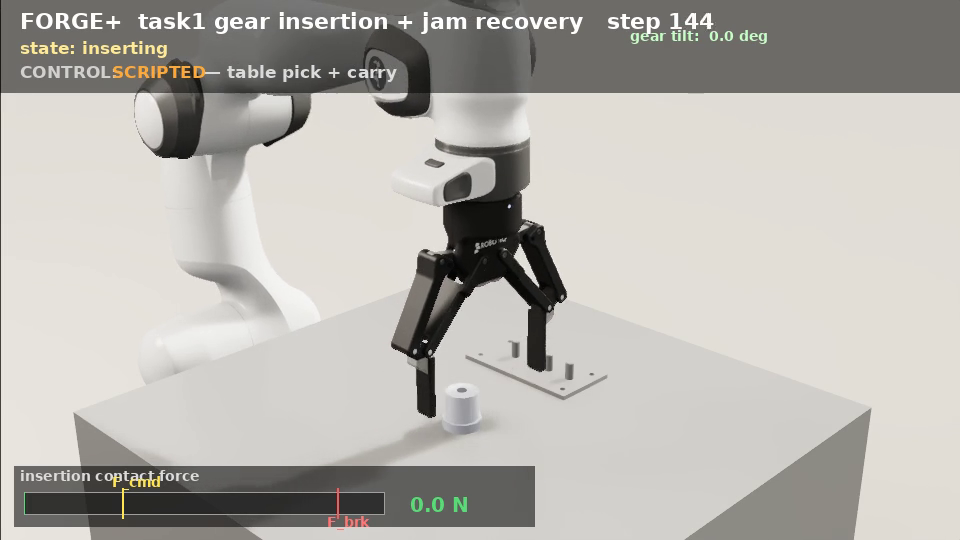}\hfill
\includegraphics[width=0.24\textwidth]{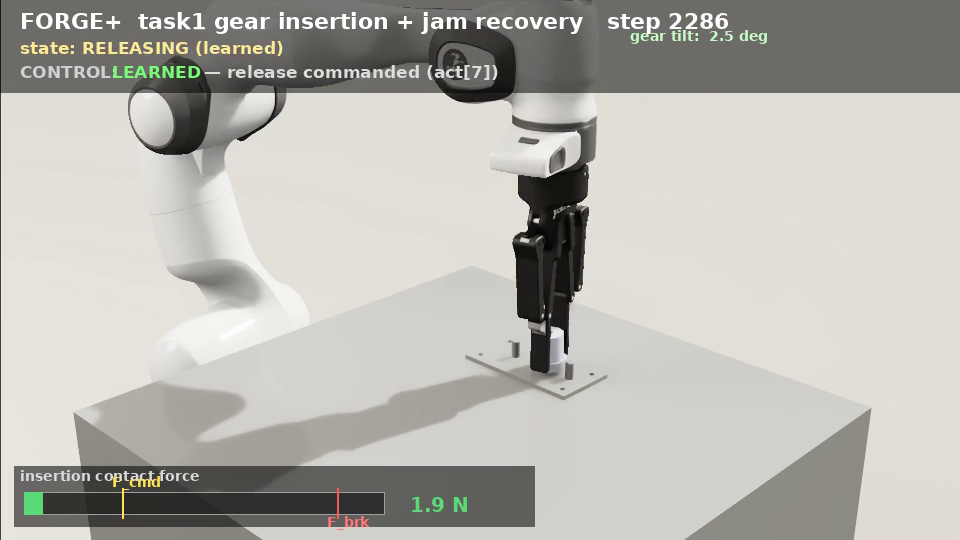}\hfill
\includegraphics[width=0.24\textwidth]{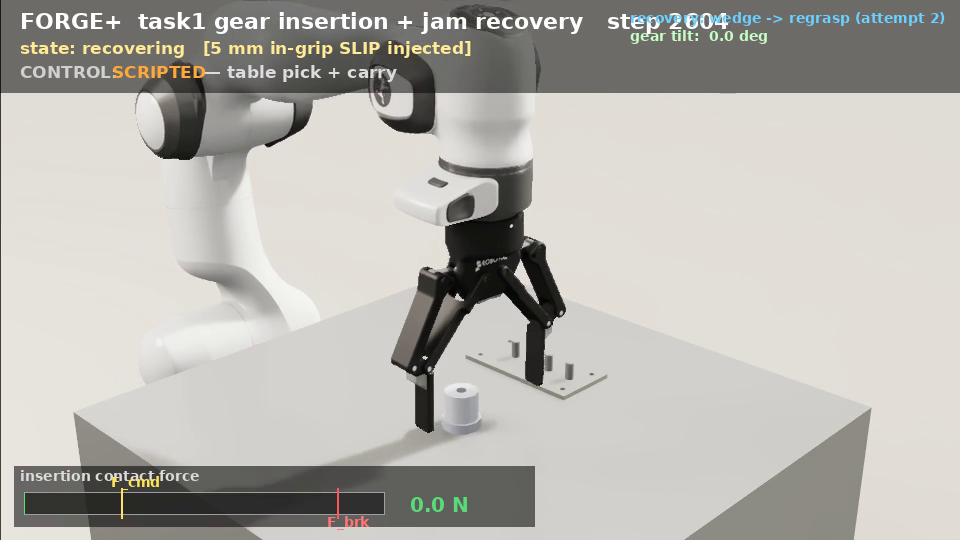}\hfill
\includegraphics[width=0.24\textwidth]{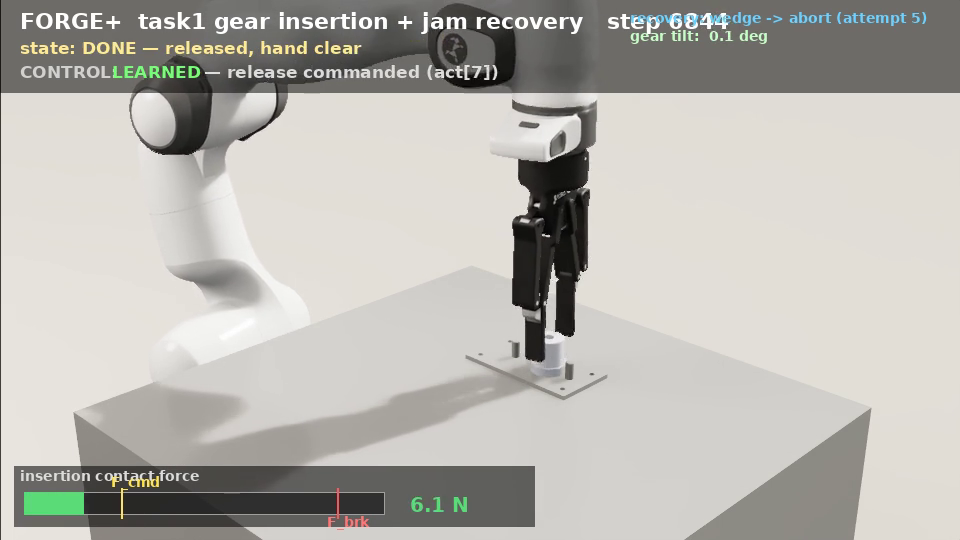}
\caption{Stills from the released 2F-140 videos (HUD labels every phase \textsc{learned}/\textsc{scripted} with a live force gauge vs.\ \Fmax{} and \Fbreak). Left to right: the scripted-staging table pick (a real friction grasp of the resting gear); the learned release firing at 1.9\,N after a clean insertion; a recovery episode mid-\code{regrasp} --- the tilted gear has been placed back on the table for a physical re-pick (LLM decision \code{wedge $\to$ regrasp}, attempt 2); the same episode ending seated, released, hand clear after two full regrasp cycles.}
\label{fig:stills}
\end{figure}

\subsection{Budget baselines: the oracle breaks half the parts}
\label{sec:exp-budget}

\begin{table}[t]
\centering
\small
\caption{Budget-setter baselines on the fragile ABS gear (Franka, deterministic mean, $\sim$200 episodes/cell, strict criterion). Margin $=$ mean$(\Fbreak) - \Fmax$. $^{*}$No-ceiling produced \emph{zero episode completions}: $\Fcmd$ conditioning far outside the training distribution degenerates behavior (no descent to contact) --- non-functional rather than safe.}
\label{tab:budget}
\resizebox{\textwidth}{!}{%
\begin{tabular}{lrrrlr}
\toprule
Budget & \Fmax & Success & Breakage & Peak force mean/p95/max (N) & Over-budget eps \\
\midrule
\textbf{Ours} (LLM, identity-only) & 10\,N & \textbf{1.000} & \textbf{0.000} & 14.9 / 16.8 / 20.2 & 0/200 \\
Oracle ($\Fbreak-5$) & $\sim$33\,N & 0.502 & 0.498 & 33.9 / 44.9 / 53.8 & 0/201 \\
Fixed global & 60\,N & 0.000 & 1.000 & 34.6 / 42.2 / 47.1 & 205/205 \\
No ceiling$^{*}$ & 120\,N & 0.000 & 0.000 & --- & 200/200 \\
\bottomrule
\end{tabular}%
}
\end{table}

Table~\ref{tab:budget} gives a stronger separation than we expected --- \textbf{ours $\gg$ oracle $>$ fixed} --- with one inversion that is a finding in itself. The oracle sets $\Fmax=\Fbreak-\varepsilon$, the ``optimal'' budget if clamp fidelity were perfect. It is not: conditioned at 33\,N, the skill's funnel-entry overshoot peaks at 45--54\,N ($\sim$1.5$\times$ the budget), past the $38\pm5$\,N breaking force in half the episodes. \textbf{Budget appropriateness must cover the overshoot distribution, not just sit under \Fbreak{}} --- the conservative identity-derived 10\,N budget absorbs the same overshoot with room to spare, seating 100\% with zero breaks. The fixed 60\,N global ceiling destroys every part (the canonical argument for per-object budgets), and the unbounded baseline is not even unsafe --- it is non-functional, because force conditioning is part of the skill's operating envelope and a far-out-of-distribution \Fcmd{} degenerates the behavior itself.

\subsection{Recovery under an in-grip slip: two grippers, five baselines}
\label{sec:exp-recovery}

\begin{table}[t]
\centering
\small
\caption{Jam-recovery sweep: fragile ABS gear, 5\,mm in-grip slip every episode, 25 episodes/cell, deterministic mean, identity-derived budget, \Fmax{} never raised. \emph{The two tables use different step caps (1600 vs.\ 700, because one 2F-140 regrasp cycle is $\sim$450 steps and the cap must fit $\sim$3 cycles) and different checkpoints, so rows are comparable within a table, not across tables.} On the Franka table the attempts counter saturated at 5 in every cell (cosmetic settle-churn re-fires) and is omitted; the 2F-140 protocol suppresses that churn.}
\label{tab:sweep}
\resizebox{\textwidth}{!}{%
\begin{tabular}{l rrrr l}
\toprule
\multicolumn{6}{c}{\textbf{Robotiq 2F-140} (1600-step cap; unified checkpoint, clean gates 256/256 both classes)} \\
\midrule
Recovery & Success & Breaks & Timeouts & Attempts & Peak force mean/max (N) \\
\midrule
\textbf{Ours} (force signature) & \textbf{40\%} (10/25) & 12\% (3) & 48\% (12) & 3.56 & 23.0 / 48.3 \\
Vision-LLM proxy & 28\% (7/25) & 12\% (3) & 60\% & 3.88 & 19.0 / 43.6 \\
Heuristic & 0\% & 0\% & 100\% & 5.00 & 12.5 / 18.0 \\
Press-harder & 0\% & 0\% & 100\% & 5.00 & 11.3 / 16.5 \\
None & 0\% & \textbf{20\% (5)} & 80\% & 0 & 10.2 / 50.5 \\
\midrule
\multicolumn{6}{c}{\textbf{Franka Panda hand} (700-step cap; clean-gate policy 200/200, 0 breaks)} \\
\midrule
Recovery & Success & Breaks & Timeouts & & Peak force mean/max (N) \\
\midrule
\textbf{Ours} (force signature) & \textbf{64\%} (16/25) & 12\% (3) & 24\% (6) & & 23.2 / 31.5 \\
Vision-LLM proxy & 32\% (8/25) & 0\% & 68\% & & 15.0 / 31.6 \\
Heuristic & 0\% & 0\% & 100\% & & 0.0 / 0.0 \\
Press-harder & 4\% (1/25) & \textbf{96\% (24)} & 0\% & & 36.1 / 44.8 \\
None & 0\% & 0\% & 100\% & & 0.0 / 0.0 \\
\bottomrule
\end{tabular}%
}
\end{table}

\begin{figure}[t]
\centering
\includegraphics[width=0.95\textwidth]{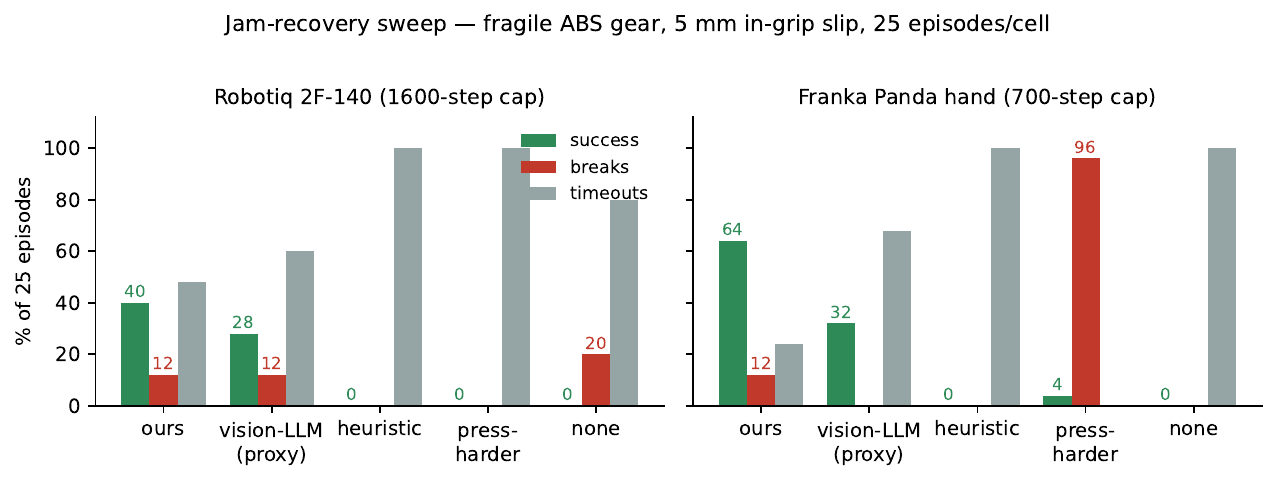}
\caption{The recovery sweep of Table~\ref{tab:sweep} as grouped bars. Note press-harder's two faces: futile on the 2F-140 (100\% timeouts, zero breaks --- the escalated ceiling never engages because the tilted bore never takes load; there, doing \emph{nothing} is the destructive cell, 20\% breaks), destructive on the Franka (96\% breaks).}
\label{fig:sweep}
\end{figure}

\textbf{The failure mechanism.} The 5\,mm slip leaves the gear \emph{tilted 7--10$^\circ$ in the grip} (position recovers within a substep; orientation never does --- probe: 4.6$^\circ\to$10.1$^\circ$ at the kick, no decay). A tilted bore cannot thread the 0.4\,mm fit band, so the deterministic policy backs off and \emph{hovers at zero contact force} --- correctly refusing a geometrically impossible insertion, and invisible to any force-threshold jam detector. A contactless-hover branch in the failure detector makes the refusal legible; the signature chain routes a \emph{recurring} hover (an anomaly that survives an arm maneuver, hence travels with the part) to \code{regrasp}, the only menu action that fixes an in-grip tilt. The re-seat restores the canonical grip (9.8$^\circ\to$0.2$^\circ$).

\textbf{Reading Table~\ref{tab:sweep}.} Only the force-signature chain ever restores an insertable grip. The hand-coded heuristic never reads the hover signature, maps every zero-force state to \code{wiggle\_search}, never regrasps: 0\%. The vision-LLM proxy --- \emph{implemented as a random menu draw}, a deliberately generous stand-in for pixels-only reasoners that cannot see in-grip force asymmetry --- succeeds exactly as often as it happens to pick \code{regrasp}: the force signature is worth 40\% vs.\ 28\% on the 2F-140 and 64\% vs.\ 32\% on the Franka (the gap is narrower on the 2F-140 because a wasted wrong-maneuver attempt burns a $\sim$450-step regrasp cycle from the step budget). We label this baseline a proxy: it bounds what any reasoner that cannot discriminate the signature could achieve by selection alone.

\textbf{Press-harder shows two faces of the same anti-pattern.} On the 2F-140 it is \emph{futile}: its only maneuver never fixes the in-grip tilt, the tilted bore never takes load, so the raised ceiling never even engages --- 0 breaks, 0 successes, 100\% timeouts at a 16.5\,N max. On the Franka it is destructive: escalating \Fmax{} $\times$1.25 per attempt destroys \textbf{24 of 25} fragile gears (peaks to 44.8\,N vs.\ \Fbreak{} $38\pm5$). Both faces refute force escalation as a recovery strategy, for different reasons; on the 2F-140 the destructive face reappears in the \emph{none} cell, where with no recovery interrupting it the policy's own descent eventually presses the tilted bore into the shaft tip (5/25 breaks at up to 50.5\,N). The jam is real, and doing nothing is worse than a wrong maneuver.

\textbf{Robust class.} On the steel gear the same chain recovers \textbf{25/25 with 0 breaks and 0 timeouts} --- but recovery episodes press 85\,N mean / 135\,N max vs.\ the 35.6/62.5 clean gate, consuming most of steel's 100\,N budget margin.

\textbf{Honest caveats.} (i) Recovery episodes press harder than clean ones on both grippers (2F-140: 23.0\,N mean / 48.3 max vs.\ 15.9 / 36.7 clean) --- that envelope exposure, not any raised ceiling, is what catches low \Fbreak{} draws and accounts for all 3 fragile breaks in each \emph{ours} cell; a gentler post-recovery descent profile is the obvious next refinement. (ii) The 48\% timeout rate in the 2F-140 \emph{ours} cell is the current frontier, not breakage: the cap fits $\sim$3 recovery cycles and detector churn can burn them. (iii) With the fully physical place-on-table regrasp and learned release enabled end-to-end, a 5-episode smoke completes 3/5 (2 breaks, 0 timeouts) --- a smoke, not a cell; the 25-episode table-flow cell is open work.

\subsection{Force economy across the arc}

Peak insertion forces on the fragile class, all with the identity-derived $\approx$10\,N budget and zero raised ceilings: pinned-staging clean gate 15.9\,N mean; with learned release 15.7\,N; full table-pick flow \textbf{5.4\,N mean / 5.8 max}; recovery episodes 23.0\,N mean / 48.3 max (envelope exposure, \S\ref{sec:exp-recovery}). The system's gentlest contact came from \emph{removing} scaffolding, and its hardest pressing comes from post-recovery seating --- the measured cost of a second chance.

\section{Limitations}
\label{sec:limitations}

\textbf{Simulation only; breakage is a scalar.} No real robot, no sim-to-real claim. Breakage is a hidden threshold on peak contact force in rigid-body physics --- no fracture, deformation, or fatigue. FORGE-style force conditioning is independently known to transfer~\citep{noseworthy2025forge}; hardware is deferred, not implied.

\textbf{The clamp bounds the command, not the contact.} Impedance overshoot reaches $\sim$1.5$\times$ the budget at funnel entry (\S\ref{sec:exp-budget}). We surface it as the clamp-fidelity metric and absorb it with conservative budgets; a contact-force kill switch is an open design choice.

\textbf{Disturbances are injected and simple.} The rim wedge and the 5\,mm slip are controllable, reproducible faults --- not a naturalistic failure distribution. Richer jams (angular catch, burrs, cross-threading) are future work.

\textbf{The vision baseline is a proxy.} We compare against a menu-sampling stand-in, not a full REFLECT/AHA pipeline with rendered frames; the comparison bounds selection-only performance rather than testing a specific vision system.

\textbf{Scripted components remain, and are labeled.} Transport staging and recovery primitives are scripted (the LLM selects; the primitive executes); the release head is staging-specific (it reads a phase one-hot) and must be recollected per flow. Several table-flow results are explicitly smokes (64-episode release gate; 3/5 regrasp cell), and the 2F-140 recovery cell's 48\% timeout rate is an open frontier.

\textbf{LLM scope.} The budget-setter reasons over provided identity text; we did not evaluate adversarial or paraphrased identities, and budget outputs are cached per class. Whether the identity-derived budget is fully gripper-invariant is measured only indirectly here (the same class budgets drive both grippers' evaluations).

\section{Conclusion}

FORGE-plus asked two questions that force-conditioned assembly skills leave open: who sets the ceiling, and what to do on failure. The answers that survive our evaluation are structural, not model-flavored. A conservative, identity-derived budget beats the oracle, because a real budget must cover the controller's overshoot distribution --- not flirt with \Fbreak{}. A recovery layer earns its keep only if it can \emph{discriminate} failures: the force signature is worth a factor of $\sim$2 over selection-by-luck on both grippers, entirely because it is the only signal that routes a recurring contactless hover to the one maneuver that fixes a tilted grip. And ``press harder,'' the natural recovery, fails in two distinct ways on two grippers --- destructive where the load path engages, futile where it does not --- while never once being the right answer. Meanwhile, the safety story never depended on the language model: the clamp did.

We are equally interested in what did not work. On-policy RL is excluded at 0.4\,mm clearance by the exploration--damage trade-off itself; the practical pipeline is expert demos, DAgger, BC, and a weight soup; a learned release decision needs an input-skip head because trained trunks provably discard the state it depends on. We release the code, the evaluation scripts, the checkpoint manifest (including the nine failed PPO lineages kept as evidence), and the labeled videos.

\bibliographystyle{unsrtnat}
\bibliography{references}

@article{noseworthy2025forge,
  title   = {{FORGE}: Force-Guided Exploration for Robust Contact-Rich Manipulation under Uncertainty},
  author  = {Noseworthy, Michael and Tang, Bingjie and Wen, Bowen and Handa, Ankur and Kessens, Chad and Roy, Nicholas and Fox, Dieter and Ramos, Fabio and Narang, Yashraj and Akinola, Iretiayo},
  journal = {IEEE Robotics and Automation Letters},
  year    = {2025},
  note    = {arXiv:2408.04587}
}

@article{huang2025tactilevla,
  title   = {Tactile-{VLA}: Unlocking Vision-Language-Action Model's Physical Knowledge for Tactile Generalization},
  author  = {Huang, Jialei and Wang, Shuo and Lin, Fanqi and Hu, Yihang and Wen, Chuan and Gao, Yang},
  journal = {arXiv preprint arXiv:2507.09160},
  year    = {2025}
}

@article{cao2026pacovla,
  title   = {{PaCo-VLA}: Passivity-Shielded Compliance Prior for Contact-Rich Vision-Language-Action Manipulation},
  author  = {Cao, Yifan and others},
  journal = {arXiv preprint arXiv:2606.00515},
  year    = {2026}
}

@inproceedings{liu2023reflect,
  title     = {{REFLECT}: Summarizing Robot Experiences for Failure Explanation and Correction},
  author    = {Liu, Zeyi and Bahety, Arpit and Song, Shuran},
  booktitle = {Conference on Robot Learning (CoRL)},
  year      = {2023},
  note      = {arXiv:2306.15724}
}

@article{guo2023doremi,
  title   = {{DoReMi}: Grounding Language Model by Detecting and Recovering from Plan-Execution Misalignment},
  author  = {Guo, Yanjiang and Wang, Yen-Jen and Zha, Lihan and Jiang, Zheyuan and Chen, Jianyu},
  journal = {arXiv preprint arXiv:2307.00329},
  year    = {2023}
}

@article{duan2024aha,
  title   = {{AHA}: A Vision-Language-Model for Detecting and Reasoning Over Failures in Robotic Manipulation},
  author  = {Duan, Jiafei and Pumacay, Wilbert and Kumar, Nishanth and Wang, Yi Ru and Tian, Shulin and Yuan, Wentao and Krishna, Ranjay and Fox, Dieter and Mandlekar, Ajay and Guo, Yijie},
  journal = {arXiv preprint arXiv:2410.00371},
  year    = {2024}
}

@inproceedings{narang2022factory,
  title     = {Factory: Fast Contact for Robotic Assembly},
  author    = {Narang, Yashraj and Storey, Kier and Akinola, Iretiayo and Macklin, Miles and Reist, Philipp and Wawrzyniak, Lukasz and Guo, Yunrong and Moravanszky, Adam and State, Gavriel and Lu, Michelle and Handa, Ankur and Fox, Dieter},
  booktitle = {Robotics: Science and Systems (RSS)},
  year      = {2022},
  note      = {arXiv:2205.03532}
}

@inproceedings{tang2023industreal,
  title     = {{IndustReal}: Transferring Contact-Rich Assembly Tasks from Simulation to Reality},
  author    = {Tang, Bingjie and Lin, Michael A. and Akinola, Iretiayo and Handa, Ankur and Sukhatme, Gaurav S. and Ramos, Fabio and Fox, Dieter and Narang, Yashraj},
  booktitle = {Robotics: Science and Systems (RSS)},
  year      = {2023},
  note      = {arXiv:2305.17110}
}

@inproceedings{tang2024automate,
  title     = {{AutoMate}: Specialist and Generalist Assembly Policies over Diverse Geometries},
  author    = {Tang, Bingjie and Akinola, Iretiayo and Xu, Jie and Wen, Bowen and Handa, Ankur and Van Wyk, Karl and Fox, Dieter and Sukhatme, Gaurav S. and Ramos, Fabio and Narang, Yashraj},
  booktitle = {Robotics: Science and Systems (RSS)},
  year      = {2024},
  note      = {arXiv:2407.08028}
}

@article{mittal2025isaaclab,
  title   = {Isaac Lab: A GPU-Accelerated Simulation Framework for Multi-Modal Robot Learning},
  author  = {Mittal, Mayank and others},
  journal = {arXiv preprint arXiv:2511.04831},
  year    = {2025}
}

@article{mittal2023orbit,
  title   = {Orbit: A Unified Simulation Framework for Interactive Robot Learning Environments},
  author  = {Mittal, Mayank and Yu, Calvin and Yu, Qinxi and Liu, Jingzhou and Rudin, Nikita and Hoeller, David and Yuan, Jia Lin and Singh, Ritvik and Guo, Yunrong and Mazhar, Hammad and Mandlekar, Ajay and Babich, Buck and State, Gavriel and Hutter, Marco and Garg, Animesh},
  journal = {IEEE Robotics and Automation Letters},
  year    = {2023},
  note    = {arXiv:2301.04195}
}

@article{murali2025graspgen,
  title   = {{GraspGen}: A Diffusion-based Framework for 6-DOF Grasping},
  author  = {Murali, Adithyavairavan and others},
  journal = {arXiv preprint arXiv:2507.13097},
  year    = {2025}
}

@article{yu2025forcevla,
  title   = {{ForceVLA}: Enhancing {VLA} Models with a Force-aware {MoE} for Contact-Rich Manipulation},
  author  = {Yu, Jiawen and others},
  journal = {arXiv preprint arXiv:2505.22159},
  year    = {2025}
}

@article{zhang2026compliantvla,
  title   = {{CompliantVLA}-adaptor: {VLM}-Guided Variable Impedance Action for Safe Contact-Rich Manipulation},
  author  = {Zhang, Wei and Huang, Ying and others},
  journal = {arXiv preprint arXiv:2601.15541},
  year    = {2026}
}

@article{schulman2017ppo,
  title   = {Proximal Policy Optimization Algorithms},
  author  = {Schulman, John and Wolski, Filip and Dhariwal, Prafulla and Radford, Alec and Klimov, Oleg},
  journal = {arXiv preprint arXiv:1707.06347},
  year    = {2017}
}

@inproceedings{ross2011dagger,
  title     = {A Reduction of Imitation Learning and Structured Prediction to No-Regret Online Learning},
  author    = {Ross, St{\'e}phane and Gordon, Geoffrey and Bagnell, Drew},
  booktitle = {International Conference on Artificial Intelligence and Statistics (AISTATS)},
  year      = {2011},
  note      = {arXiv:1011.0686}
}

@inproceedings{wortsman2022soups,
  title     = {Model Soups: Averaging Weights of Multiple Fine-Tuned Models Improves Accuracy Without Increasing Inference Time},
  author    = {Wortsman, Mitchell and Ilharco, Gabriel and Gadre, Samir Ya and Roelofs, Rebecca and Gontijo-Lopes, Raphael and Morcos, Ari S. and Namkoong, Hongseok and Farhadi, Ali and Carmon, Yair and Kornblith, Simon and Schmidt, Ludwig},
  booktitle = {International Conference on Machine Learning (ICML)},
  year      = {2022},
  note      = {arXiv:2203.05482}
}

\end{document}